\pdfoutput=1
\documentclass[letterpaper, 10 pt, conference]{ieeeconf}  

\IEEEoverridecommandlockouts                              

\usepackage{cite}
\usepackage{amsmath,amssymb,amsfonts}
\usepackage{algorithmic}
\usepackage{graphicx}
\usepackage{textcomp}
\usepackage{xcolor}
\usepackage{subcaption}
\usepackage{multirow}
\usepackage{booktabs}
\usepackage{colortbl}
\usepackage{anyfontsize}
\usepackage{t1enc}
\usepackage{caption}
\usepackage{hyperref}
\definecolor{tabcolor}{RGB}{235,235,235}
\usepackage[misc]{ifsym}

\title{\LARGE \bf
Kinematic-aware Prompting for Generalizable Articulated Object Manipulation with LLMs
}

\author{Wenke Xia$^{1,2,\ast,\dagger}$,  Dong Wang$^{2,\ast}$, Xincheng Pang$^{1}$, Zhigang Wang$^{2}$, Bin Zhao$^{2,3}$, Di Hu$^{1,\ddagger}$, Xuelong Li$^{2,4,\ddagger}$
\thanks{$^{1}$Gaoling School of Artificial Intelligence, Renmin University of China}%
\thanks{$^{2}$Shanghai Artificial Intelligence Laboratory}%
\thanks{$^{3}$Northwestern Polytechnical University}%
\thanks{$^{4}$Institute of Artificial Intelligence, China Telecom Corp Ltd}%
\thanks{$^{\ast}$Equal contribution,$^{\dagger}$Work is done during internship at Shanghai Artificial Intelligence Laboratory, $^{\ddagger}$Corresponding author}%
}

\begin{document}

\maketitle
\thispagestyle{empty}
\pagestyle{empty}

\begin{abstract}

Generalizable articulated object manipulation is essential for home-assistant robots. Recent efforts focus on imitation learning from demonstrations or reinforcement learning in simulation, however, due to the prohibitive costs of real-world data collection and precise object simulation, it still remains challenging for these works to achieve broad adaptability across diverse articulated objects. 
Recently, many works have tried to utilize the strong in-context learning ability of Large Language Models (LLMs) to achieve generalizable robotic manipulation, but most of these researches focus on high-level task planning, sidelining low-level robotic control.
In this work, building on the idea that the kinematic structure of the object determines how we can manipulate it, we propose a kinematic-aware prompting framework that prompts LLMs with kinematic knowledge of objects to generate low-level motion trajectory waypoints, supporting various object manipulation.
To effectively prompt LLMs with the kinematic structure of different objects, we design a unified kinematic knowledge parser, which represents various articulated objects as a unified textual description containing kinematic joints and contact location.
Building upon this unified description, a kinematic-aware planner model is proposed to generate precise 3D manipulation waypoints via a designed kinematic-aware chain-of-thoughts prompting method.
Our evaluation spanned 48 instances across 16 distinct categories, revealing that our framework not only outperforms traditional methods on 8 seen categories but also shows a powerful zero-shot capability for 8 unseen articulated object categories with only 17 demonstrations.
Moreover, the real-world experiments on 7 different object categories prove our framework's adaptability in practical scenarios. Code is released at \href{https://github.com/GeWu-Lab/LLM_articulated_object_manipulation}{https://github.com/GeWu-Lab/LLM\_articulated\_object\_manipulation}.

\end{abstract}

\section{INTRODUCTION}

\begin{figure}[t]
    \centering
    \includegraphics[width=0.47\textwidth]{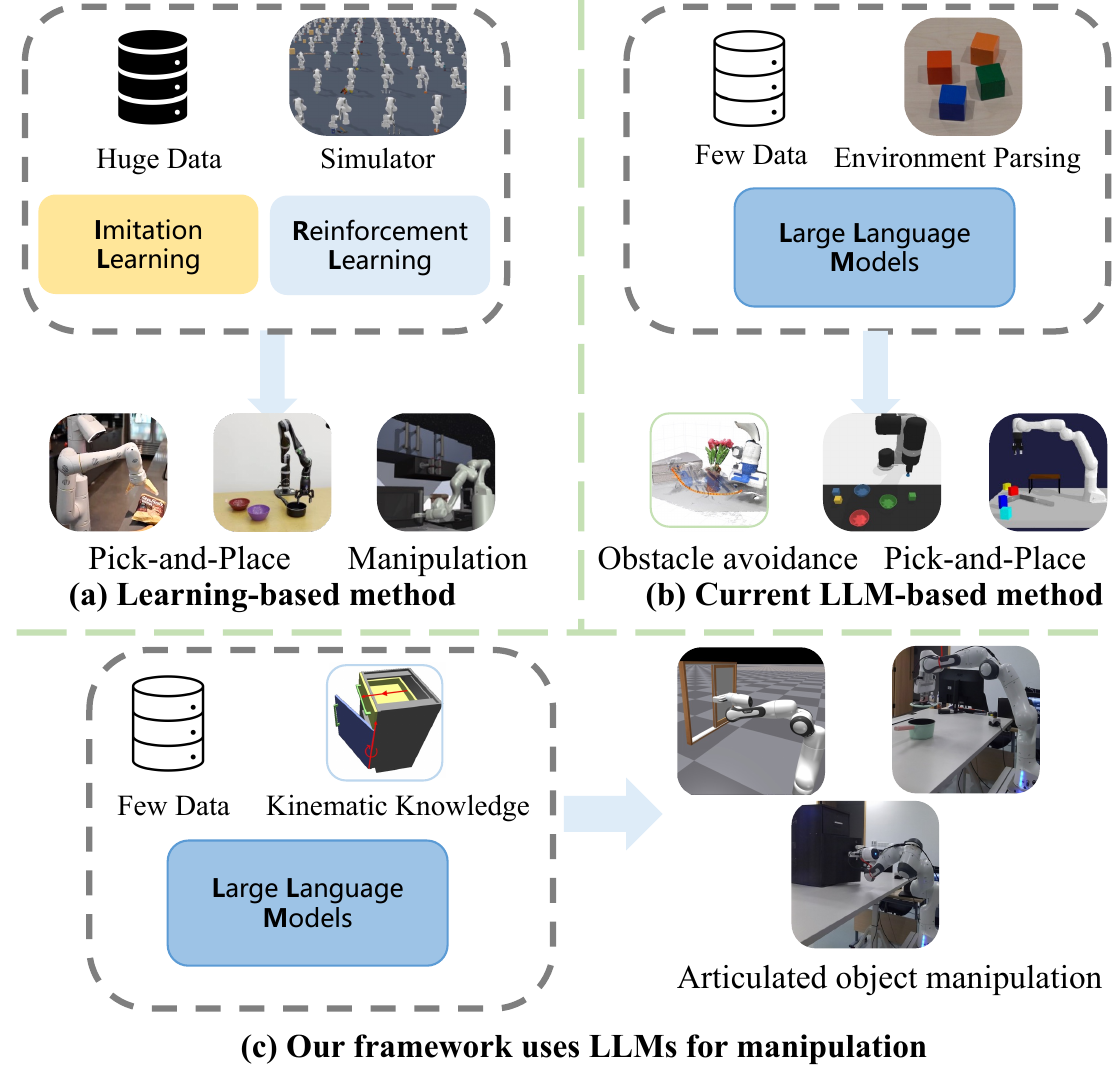}
    
    \caption{ As depicted in (a), traditional learning-based methods rely on vast datasets for broad manipulation tasks. Recent studies in (b) harness LLMs to reduce data reliance, but primarily apply to elementary challenges like obstacle avoidance, and pick-and-place. In contrast, our framework, highlighted in (c), achieves zero-shot articulated object manipulation with the kinematic-aware prompting method.}
    \label{fig:intro}
    \vspace{-2em}
\end{figure}

Generalizable articulated object manipulation is imperative for building intelligent and multi-functional robots.
However, due to the considerable heterogeneity in the kinematic structures of objects, the manipulation policy might vary drastically across different object instances and categories. 
To ensure consistent performance in automated tasks within intricate real-world scenarios, prior works on generalizable object manipulation have been devoted to imitation learning from demonstrations~\cite{jia2023chain,shridhar2023perceiver} and reinforcement learning in simulation~\cite{geng2023rlafford,gu2023maniskill2}. As shown in Figure~\ref{fig:intro}(a), these approaches consistently require substantial amounts of robotic data. To bolster the manipulation generalization, recent works~\cite{brohan2022rt,brohan2023rt} have made notable advancements in curating extensive robotic datasets. However, the diversity of scenarios covered in these datasets is still limited. To achieve generalization across various scenarios, there would be a prohibitive cost of accumulating more data.
Thus, contemporary studies strived to mitigate this excessive data dependency through strategies such as efficient architectures~\cite{shridhar2023perceiver,mees2022matters}, redefined training objectives~\cite{jia2023chain,shi2023waypoint}, and data augmentation techniques~\cite{roboagent,zhou2023nerf}. Nevertheless, the generalizability of these solutions remains limited, especially in novel scenarios that contain unseen object states.

In pursuit of developing a generalizable object manipulation policy with reduced data reliance, recent works incorporated rich world knowledge within LLMs to promote policy learning~\cite{liu2023softgpt,tang2023graspgpt}.
Drawing insights from the reasoning capabilities of LLMs~\cite{wei2022chain,wei2022emergent}, 
as shown in Figure~\ref{fig:intro}(b), some contemporary studies~\cite{mirchandani2023large,mandi2023roco} successfully used LLMs to predict reasonable action sequences for object manipulation with a parsed textual representation of environments. However, such a coarse representation limits these works to simple tasks like pick-place and rearrangement planning. Although Huang~et~al.~\cite{huang2023voxposer} introduced a 3D voxel map to capture intricate environment details and tackle challenges like obstacle avoidance, there is still a notable gap in handling sophisticated articulated object manipulation tasks.

~\label{section:intro}
In this work, we delve into the problem of harnessing LLMs for generalizable articulated object manipulation, recognizing that the rich world knowledge inherent in LLMs is adept at providing reasonable manipulation understanding of various articulated objects.
For instance, when presented with the instruction ``open the cabinet", LLMs can provide a systematic approach: 1) Locate the handle or knob, 2) Determine the hinge direction, and 3) Either push or pull based on the hinge type.
However, to fully leverage the rich world knowledge within LLMs for precise articulated object manipulation, we still confront the critical challenge of converting these abstract manipulation commonsense into precise low-level robotic control.  

To tackle the aforementioned challenge, we propose a kinematic-aware prompting framework. This framework is designed to extract the kinematic knowledge of various objects and prompt LLMs to generate low-level motion trajectory waypoints for object manipulations as shown in Figure~\ref{fig:intro}(c). The idea behind this method is that the kinematic structure of an object determines how we can manipulate it.
Therefore, we first propose a unified kinematic knowledge parser, which represents the various articulated objects as a unified textual description with the kinematic joints and contact location. Harnessing this unified description, a kinematic-aware planner is proposed to generate precise 3D manipulation waypoints for articulated object manipulation via a kinematic-aware chain-of-thought prompting. Concretely, it initially prompts LLMs to generate an abstract textual manipulation sequence under the kinematic structure guidance. Subsequently, it takes the generated kinematic-guided textual manipulation sequence as inputs, and outputs 3D manipulation trajectory waypoints via in-context learning for precise robotic control. With this kinematic-aware hierarchical prompting, our framework can effectively utilize LLMs to understand various object kinematic structures to achieve generalizable articulated object manipulation.

To validate the efficacy of our framework, we conduct exhaustive experiments on 48 objects across 16 categories in Isaac Gym~\cite{makoviychuk2021isaac} simulator and extend our method to real-world scenarios. The results prove that our framework could generalize across seen/unseen object instances and categories in a zero-shot context. Moreover, the real-world experiments prove our framework's ability to extend its generalization to practical scenarios.

The main contributions of our work are as follows:
\begin{itemize}
    \item We propose the Kinematic-aware prompt framework, aiming for generalizable articulated object manipulation across novel instances and categories with minimal robotic data requirements.
    \item We design the unified kinematic knowledge parser and kinematic-aware planner components, utilizing the kinematic knowledge to prompt LLMs to generate precise 3D manipulation trajectory waypoints.
    \item We evaluate our method on 48 instances across 16 categories. The results prove our framework exhibits zero-shot ability for articulated object manipulation. The real-world experiments also prove our framework's generalization to practical scenarios.
\end{itemize}

\section{Related Works}
\subsection{Policy Learning for object manipulation}

Object manipulation policy learning methods have primarily focused on imitation learning from demonstrations~\cite{jia2023chain,brohan2022rt,zhang2023affordance} and reinforcement learning~\cite{geng2023rlafford} in simulations. To devise a practical manipulation policy for specific scenarios, these methods~\cite{brohan2022rt,brohan2023rt,nair2022learning,mees2022matters} often rely on numerous demonstrations or episodes.
To reduce the cost of data collection, some researchers~\cite{rao2020rl, mandi2022cacti, chen2023genaug, roboagent} employ generative models to augment the limited robotic data for robotic policy training. 
In addition to innovations in dataset augmentation, recent works~\cite{jia2023chain,shridhar2023perceiver} seek to bolster the learning efficacy of models. Leveraging the transformer architecture, numerous studies have showcased efficient policy learning in a limited demonstration dataset~\cite{chen2021decision,shridhar2023perceiver}. 
Moreover, Jia~et~al.~\cite{jia2023chain} incorporates the idea of hierarchical reinforcement learning with imitation learning for generalizable policy learning.
However, these studies still face challenges in unseen scenarios when training with a limited dataset.
In this work, we propose a kinematic-aware prompting framework that guides LLMs to generate low-level motion trajectory waypoints with the object kinematic knowledge, thereby facilitating a more generalized approach to articulated object manipulation with minimal reliance on robotic demonstrations.

\begin{figure*}[t]
    \centering
    \includegraphics[width=0.90\textwidth]{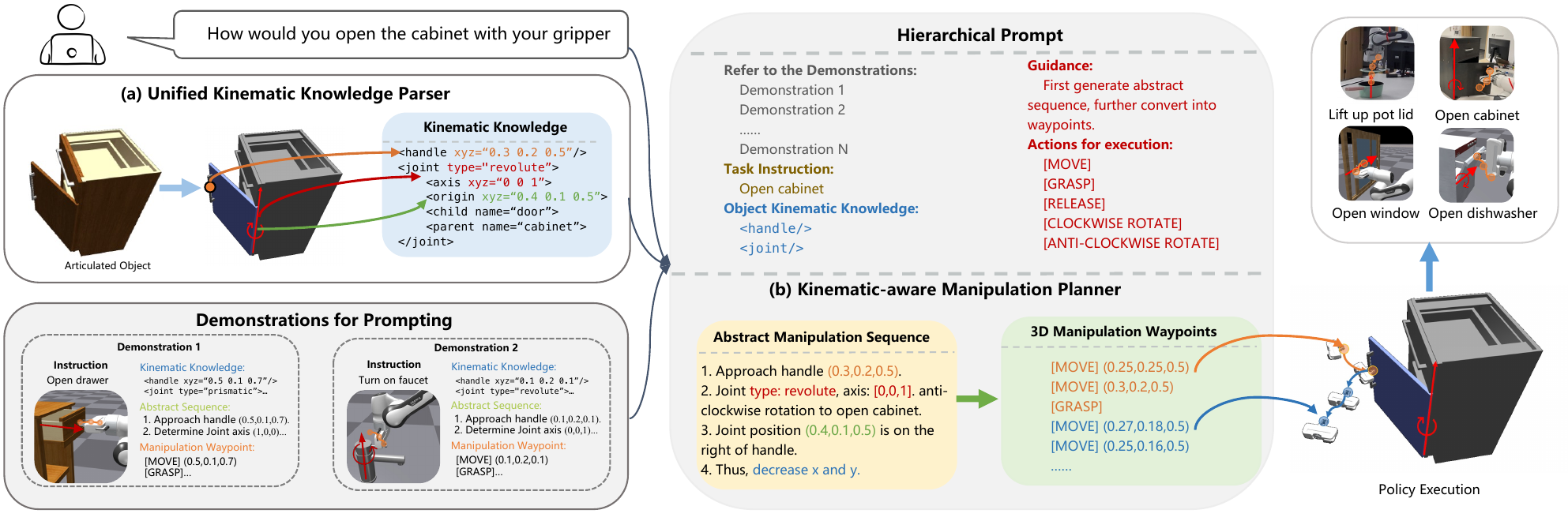}
    
    \caption{We first propose the Unified Kinematic Knowledge Parser component to grasp the object's kinematic structure as a kinematic knowledge description for LLMs as shown in (a). Based on the description, we construct a kinematic-aware hierarchical prompt, which is used in the Kinematic-aware Manipulation Planner component to guide LLMs to generate an abstract textual manipulation sequence, and 3D manipulation waypoints for generalizable articulated object manipulation in (b). Distinct colors assigned to numbers represent the properties of the different kinematic structure components.}
    \label{fig:pipeline}
\end{figure*}

\subsection{Large Language Models for Robotics}
Motivated by the rich world knowledge exhibited by LLMs, recent literature has explored the integration of LLMs with robotics across various domains~\cite{ren2023robots,huang2022inner, zhou2023navgpt}. To enable robots to adapt to complex real-world scenarios, many works~\cite{ahn2022can,wu2023tidybot,chen2023open,jansen2020visually,wu2023embodied,zhu2023ghost,yuan2023plan4mc,driess2023palm,vemprala2023chatgpt,singh2023progprompt,liang2023code} focus on task planning and the decomposition of complex instructions. Although these studies exhibit superior planning ability to decompose complex unseen instructions into subgoals, inevitably, they still depend on a pre-trained skill library for the fulfillment of subgoals. However, this dependence poses challenges due to the scarcity of extensive robotic datasets for learning various skills. To improve this skill acquisition process, some researchers~\cite{yu2023language, di2023towards} employ LLMs for reward designing. Moreover, Mirchandani~et~al.~\cite{mirchandani2023large} encodes actions into separate tokens and leverages LLMs to generate corresponding token sequences for robotic control through in-context learning. While, Huang~et~al.~\cite{huang2023voxposer} and Zhao~et~al.~\cite{mandi2023roco} construct the environment information to prompt LLMs to produce action sequences for manipulation. However, these works primarily focus on elementary manipulation tasks such as obstacle avoidance and pick-and-place tasks, exhibiting shortcomings in the manipulation of complex articulated objects.
To harness the full potential of LLMs for articulated object manipulation, we extract object kinematic knowledge to prompt LLMs to generate precise 3D manipulation waypoints, and achieve zero-shot manipulation for articulated objects across novel instances and categories.

\section{Method}
We aim to solve generalizable articulated object manipulation problems that require kinematic and geometric reasoning of objects to generate precise manipulation policy. As shown in Figure~\ref{fig:pipeline}, the proposed kinematic-aware prompting framework is composed of two modules: Unified Kinematic Knowledge Parser and Kinematic-aware Manipulation Planner. Given the manipulation instruction $\mathbf{I}$, the unified kinematic knowledge parser (Sec.~\ref{subsection:parser}) extracts the kinematic structure of various articulated objects $\mathbf{O}$ and formats it as a unified kinematic knowledge description $\mathbf{K}$. Then, the kinematic-aware manipulation planner (Sec.~\ref{subsection:planner}) incorporates this kinematic description with few manipulation demonstrations to prompt LLMs to generate a sequence of 3D manipulation waypoints $\mathbf{W}=\{w_1,w_2,..,w_n\}$ for manipulation. Finally, the manipulation policy is executed via waypoints following a traditional motion planning algorithm. 


\subsection{Unified Kinematic Knowledge Parser}
\label{subsection:parser}
The manipulation policy of an articulated object is mostly determined by its kinematic structure~\cite{geng2023gapartnet,liu2022akb}, and the strong complex reasoning capability of LLMs presents a promising pathway to general kinematic structure understanding. For manipulating objects with different kinematic structures, a unified and effective kinematic knowledge description is essential for LLMs to understand various articulated objects and subsequently generate manipulation policy. Thus, as depicted in Figure~\ref{fig:pipeline}(a), we propose the unified kinematic knowledge parser component to represent the articulated object as a unified textual kinematic description for kinematic structure understanding. Concretely, the unified kinematic knowledge parser consists of two distinct steps. 

First, we detect and segment the geometric and kinematic structures of a given articulated object via an off-the-shelf model. In the simulator, we could directly extract this information. While, in the real world, we rely on existing perception models~\cite{geng2023gapartnet,kirillov2023segment} for joint estimation to obtain this information. The output of this step consists of geometric-linked parts, kinematic joints between parts, and a contact point for the manipulation. These part segments, joint properties, and the contact point represent the kinematic structure of this object and determine how to manipulate it.

Inspired by the remarkable aptitude of LLMs in parsing structured textual data~\cite{liang2023code}, ultimately, we translate the kinematic joint structure and contact location into a unified structured \emph{.xml} format.
Using the proposed unified kinematic knowledge parser, the complex kinematic structure of different articulated objects can be easily understood by LLMs via this code-like textual kinematic knowledge description $\mathbf{K}$.

\begin{figure*}[t]
    \centering
    \includegraphics[width=0.95\textwidth]{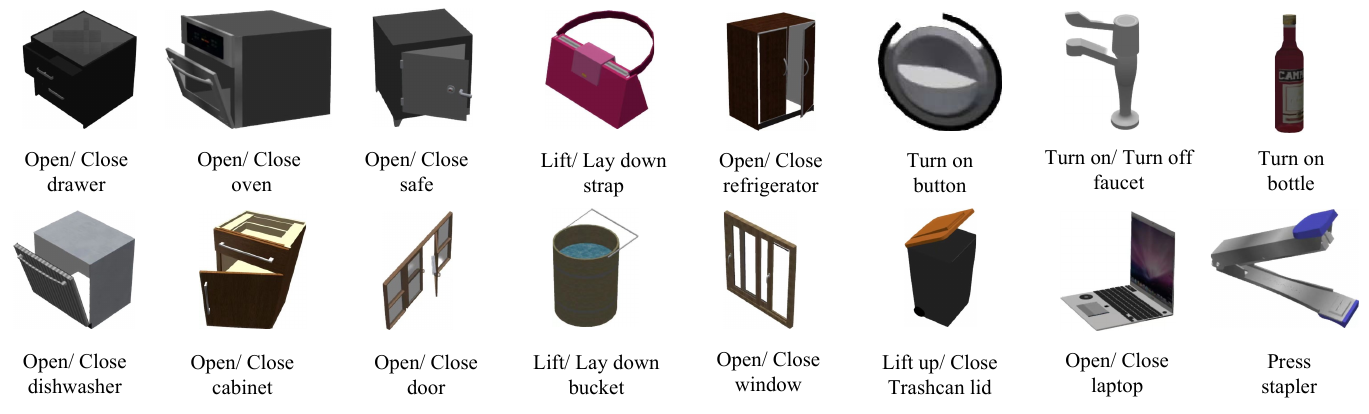}
    
    \caption{The illustration of the articulated objects used in our experiments. Each of these entities corresponds to either a singular or a pair of manipulation instructions.}
    \label{fig:assets}
    \vspace{-1em}
\end{figure*}

\subsection{Kinematic-aware Manipulation Planner}
\label{subsection:planner}


    

To enable LLMs to generate precise 3D waypoints for articulated object manipulation, we propose a hierarchical prompting method named kinematic-aware planner component, to prompt the LLMs with the unified kinematic knowledge description $\mathbf{K}$ and the manipulation instruction $\mathbf{I}$. Concretely, the hierarchical prompting method is composed of manipulation sequence planning and manipulation waypoints generation, achieving abstract textual manipulation planning to precise manipulation waypoints conversion via chain-of-thoughts prompting.

\textbf{Manipulation Sequence Planning.} As shown in Figure~\ref{fig:pipeline}, given an articulated object $\mathbf{O}$, a manipulation command specified by textual instruction $\mathbf{I}$, and its kinematic description $\textbf{K}$ from unified kinematic knowledge parser. First, the manipulation instruction $\mathbf{I}$ and kinematic description $\textbf{K}$ are concatenated as the header of the prompting. Then, we prompt LLMs with this concatenated text to produce an abstract textual manipulation sequence $\mathbf{A}$.  

As demonstrated in the textual manipulation sequence of Figure~\ref{fig:pipeline}(b), our hierarchical prompt first promotes LLMs to generate an abstract textual manipulation sequence. To generate a kinematic-feasible manipulation sequence, we prompt the LLMs to pay attention to the referred kinematic components in kinematic representation $\textbf{K}$ such as handle and joint.
To align the abstract manipulation plan with the concrete object kinematic knowledge, we make LLMs copy the corresponding properties (\emph{e.g.,} coordinate and joint orientation) of referred kinematic components into the textual manipulation sequence. 
With this explicit alignment between kinematic knowledge and abstract manipulation sequences, we could incorporate the rich world knowledge of LLMs into 3D spatial reasoning for articulated object manipulation.
\textbf{Manipulation Waypoints Generation.}
Following the generated textual manipulation sequence, the kinematic-aware planner then produces a sequence of 3D waypoints for precise robotic control. To apply the generated waypoints for various manipulation tasks, we provide five basic actions for LLMs to control the end-effector as follows:
(1) move: move the gripper to the target position. (2) grasp: close the gripper. (3) release: open the gripper. (4) clockwise rotate: clockwise rotate the gripper by 30 degrees. (5) anti-clockwise rotate: anti-clockwise rotate the gripper by 30 degrees. 
As shown in Figure~\ref{fig:pipeline}(b), LLMs could understand the 3D spatial information and generate waypoints with the formatted commands referring to the textual manipulation sequence via in-context learning. 
Further, utilizing an operational space controller, we can transform the output of the LLMs into low-level robotic control actions. 
With these explicit prompting between abstract manipulation sequences to precise 3D manipulation waypoints, the LLMs show a strong 3D spatial reasoning capability to generate reasonable 3D waypoints to manipulate various articulated objects.

\begin{table*}[h]
\centering
\resizebox{0.9\textwidth}{!}{
\begin{tabular}{ccccccccc}
\toprule
                & \multicolumn{8}{c}{Seen Categories with Unseen Instances and Poses}            \\ \midrule
Methods           & drawer & oven & safe  & strap & refrigerator & button & faucet & bottle \\ \midrule
LLM2Skill         & 100\%/ 100\%   & 66.6\%/ 100\%  & 33.3\%/ 66.6\%   & 0\%/ 0\% & 66.6\%/ 33.3\% & 100\% & 33.3\%/ 33.3\% &    100\%      \\
LLM2Waypoints    & 100\%/ 100\% & 100\%/ 100\% & 100\%/ 33.3\%  & 33.3\%/ 33.3\% & 100\%/ 100\%     & 100\% & 33.3\%/ 33.3\% &  100\%  \\
BC  & 33.3\%/ 33.3\%  &  33.3\%/ 33.3\%  & 33.3\%/ 66.6\%  & 33.3\%/ 66.6\%  & 33.3\%/ 66.6\% & 100\% & 33.3\%/ 0\% & 33.3\%  \\
\midrule

Ours&   100\%/ 100\% & 100\%/ 100\%  &  100\%/ 100\% & 100\%/ 100\% & 66.6\%/ 100\% & 100\% &  66.6\%/ 66.6\% &  66.6\%   \\

\bottomrule
\end{tabular}}
    \caption{Evaluation results on seen categories objects. We use $/$ to differentiate the model's performance on different manipulation commands for the same object.}
    \label{tab:unseen_instance}
\end{table*}

\begin{table*}[h]
\centering
\resizebox{0.9\textwidth}{!}{
\begin{tabular}{ccccccccc}
\toprule
& \multicolumn{8}{c}{Unseen Categories with different instances and poses}\\ \midrule
Methods   & dishwasher      & cabinet   & door & bucket & window& trashcan& laptop &stapler \\ \midrule
LLM2Skill  & 33.3\%/ 33.3\% & 0\%/ 33.3\% & 0\%/ 0\%  & 0\%/0\% & 33.3\%/ 0\%& 0\% / 0\% 
& 0\%/ 0\% 
&0\% 
\\
LLM2Waypoints   & 33.3\%/ 0\% & 33.3\%/ 33.3\% & 0\%/ 33.3\% & 33.3\%/ 0\% & 0\%/ 0\% & 33.3\%/ 0\% 
& 0\%/ 0\% 
&33.3\% 
\\
BC  &  33.3\%/ 33.3\% & 0\%/33.3\% & 0\%/ 33.3\% & 0\%/ 0\% & 0\%/ 0\%& 0\%/ 33.3\% & 0\%/ 33.3\% &100\% \\
\midrule
Ours & 66.6\%/ 100\% & 66.6\%/ 66.6\%  & 66.6\%/ 66.6\% & 66.6\%/ 100\%  & 66.6\%/ 66.6\%& 100\%/ 33.3\%  & 66.6\%/ 100\% &100\% \\
\bottomrule
\end{tabular}}
    \caption{Evaluation results on unseen categories objects. We use $/$ to differentiate the model's performance on different manipulation commands for the same object.}
    \label{tab:unseen_categories}
    \vspace{-1em}
\end{table*}

\section{Experiments}

\subsection{Experiment Setting}

In this work, we propose the kinematic-aware prompting method for zero-shot articulated object manipulation. To comprehensively evaluate the generalization capability of our framework, we first conduct experiments within the Isaac Gym simulator~\cite{makoviychuk2021isaac}, utilizing 48 distinct object instances across 16 types of articulated objects from the PartNet-Mobility dataset~\cite{Mo_2019_CVPR}. As shown in Figure~\ref{fig:assets}, our evaluation dataset contains a broad spectrum of commonplace articulated objects, which covers the diversity of manipulation policies and articulated structures. To enhance the scale of our evaluation data, we devised two opposite instructions for many object categories, like open/close window, open/close oven, lift/lay down bucket, turn on button, etc. In experiments, we provide the performance for all these manipulation instructions, and the order is consistent with Figure~\ref{fig:assets}.

To guide LLMs in generating 3D manipulation waypoints with the object kinematic knowledge, we collect 17 3D manipulation waypoints demonstrations across 8 distinct object categories for in-context learning in the kinematic-aware manipulation planner module. These demonstrations cover open/close drawers, open/close ovens, open/close safes, lift/ lay down straps, open/close refrigerators, turn on buttons, turn on/turn off faucets, and turn on bottles. 
To further comprehensively measure the performance of different methods, we divide the dataset into two subsets. The first subset comprises objects from eight categories of provided demonstrations, but with diverse poses and instances. The second is devoted to object categories unseen from the demonstrations, which is more challenging for the LLMs' reasoning capability and commonsense. During the evaluation, we randomly place the object in a reachable position for the robotic arm.

In each experiment, we employ a simple operational space controller to follow generated 3D waypoints to manipulate objects in simulation, and each object category is evaluated thrice by randomly selected object instances, constrained by the API cost limitation. The results are reported by the Average Success Rate (ASR) of manipulation for assessing overall performance.

\subsection{Comparison Experiments}~\label{section:ablation}
We compare our method with other approaches:
\begin{itemize}
    \item \textbf{LLM2Skill:} We implement LLM2Skill baseline as a variant of Code as Policy~\cite{liang2023code}. We predefined 18 action primitives that could finish both the demonstrated and novel instructions. Here, LLMs would determine the suitable action primitive given the detailed object kinematic knowledge.
    \item \textbf{LLM2Waypoints:} We implement this method as a naive attempt to directly output manipulation waypoints for articulated object manipulation without considering the kinematic knowledge.
    \item \textbf{Behavior Cloning.} We train a language-conditioned behavior cloning algorithm on the demonstrations, leveraging the structure of Decision Transformer~\cite{chen2021decision}.
\end{itemize}

We systematically evaluate these methods on the divided two subsets, and the results are as follows:

\textbf{Results on unseen instances and pose:} We evaluate the methods on seen categories, but with different poses and instances. As illustrated in Table~\ref{tab:unseen_instance}, most LLM-based methods were able to exhibit considerable performance on these familiar categories, drawing strength from their robust in-context learning capabilities and a wealth of inherent commonsense knowledge.
Conversely, it is challenging for learning-based methods to generalize to previously unseen instances, primarily due to the lack of demonstration data.

Notably, we discerned that similar manipulation policies might be applicable across diverse instances, allowing the LLM2Skill method to demonstrate appreciable performance on these relatively easy categories such as drawers, and buttons. However, when faced the variations within the kinematic structures, LLM2Skill fails due to the inability to craft novel trajectories for different kinematic structures like lifting straps and turning on faucets. Meanwhile, our method could still generalize to these object instances with a more flexible manipulation policy benefiting from the comprehension of kinematic knowledge.



    


\textbf{Results on unseen categories:} 
We extended our evaluation to objects within unseen categories.
As shown in Table~\ref{tab:unseen_categories}, LLMs could easily generalize to prismatic articulated objects like kitchen pots, given that the manipulation trajectory is a simple straightforward linear path.
Conversely, when manipulating revolute articulated objects, these baseline models exhibit a notable decline in the average success rate. This decline is due to that objects with revolute joints require more complex trajectories to manipulate. For example, when attempting to open a door, the generated trajectory must account for both the radius and angle to align with the object's kinematic structure. Otherwise, it would get stuck due to force constraint issues. Nonetheless, leveraging the comprehension of the object's kinematic knowledge provided by our unified kinematic knowledge parser component, our method is adept at manipulating these revolute objects.

Despite equipping the LLM2Skill method with action candidates tailored to these novel categories, it still fails due to the lack of 3D reasoning capabilities essential for versatile articulated object manipulation.

Although the behavior cloning method shows some success on a few tasks, we find that these successes are the result of random movements in the environment. Such unpredictability is not usable for real-world applications, given the inherent risks. However, by representing the action as a sequence of manipulation waypoints, our method could combine with traditional motion planners to ensure safety and applicability in real-world scenarios.

\begin{table*}[h] 
\centering

\resizebox{0.9\textwidth}{!}{
\begin{tabular}{cccccccccccc}
\toprule
 & \multicolumn{6}{c}{Unseen Categories with different instances and poses}    \\ \midrule
Method & dishwasher      & cabinet   & door & bucket & window & trashcan & laptop & stapler  \\ \midrule
  W/o. kinematic-planer planner  & 33.3\%/ 66.6\% & 0\%/ 33.3\% & 33.3\%/ 33.3\% & 33.3\%/ 33.3\%  & 33.3\%/ 0\% & 0\%/ 0\% & 0\%/ 0\% & 0\% \\

 W/o. waypoints & 33.3\%/ 66.6\% & 0\%/ 0\% & 0\%/ 33.3\%  & 0\%/33.3\%  & 33.3\%/ 33.3\% & 66.6\% / 0\% & 33.3\%/ 33.3\% & 66.6\%   \\


Ours &  66.6\%/ 100\% & 33.3\%/ 66.6\%  & 66.6\%/ 66.6\% & 66.6\%/ 100\% & 66.6\%/ 66.6\% & 100\%/ 33.3\%  & 66.6\%/ 100\% & 100\%  \\
\bottomrule
\end{tabular}}
    \caption{The ablation study results. We use $/$ to differentiate the model's performance on different manipulation commands for the same object.}
    \label{tab:ablation}

\end{table*}








\subsection{Ablation Experiments}

\begin{table}[t]\large
\centering
\resizebox{0.5\textwidth}{!}{
\begin{tabular}{c|ccccc}
\toprule
                & \multicolumn{2}{c|}{Unseen Instance}    & \multicolumn{3}{c}{Unseen Categories}        \\ \midrule
           Method & safe & refrigerator & dishwasher & cabinet & bucket  \\ \midrule
GPT-4        & 66.6\%/ 66.6\%  & 66.6\%/ 100\% & 66.6\%/ 100\% & 66.6\%/ 66.6\% & 66.6\%/ 100\%  \\
\midrule

GPT-3.5-turbo &  0\%/ 0\% & 33.3\%/ 33.3\% & 0\%/ 0\% & 33.3\%/ 33.3\% & 0\%/ 0\% \\

\bottomrule
\end{tabular}}
    \caption{Abalation study on different GPT models with 7 demonstrations as prompt.}
    \label{tab:gpt}
    \vspace{1em}
\end{table}

In the ablation experiments, we evaluate the effectiveness of our kinematic-aware prompt and waypoints generation method. Concretely, the model without the kinematic-aware planner component would directly utilize the unified kinematic description to generate manipulation waypoints without the hierarchical prompt. The model without waypoints would use the predefined action list as LLM2Skill but with our kinematic-aware prompting framework.
We follow the setting in comparison experiments and demonstrate the results on unseen categories. 

As shown in Table~\ref{tab:ablation}, the method with both components achieves the best performance, which proves the effectiveness of each component. Concretely, in experiments, we find that the model without the kinematic-aware planner is more likely to generate trajectories that do not correspond to the task instructions. However, the model without waypoint generation could generate reasonable manipulation sequences but fail in manipulating complex objects such as laptop and bucket, due to the lack of flexibility. Combining with both these modules, our method could fully understand the kinematic structure of objects for generalizable articulated object manipulation.


We compare the performance of different Large Language Models. Due to the token limitation of GPT3.5, we only provide 9 demonstrations across 4 categories, which contain open/close safe, lift/lay down strap, open/close refrigerator, and open/close oven. As shown in Table~\ref{tab:gpt}, we observe that GPT-4 is better at following the prompt to reason the spatial information, while GPT-3.5-turbo always fails to understand the relationship between kinematic knowledge and manipulation waypoints, and provide wrong manipulation waypoints.

\begin{figure}[t]
    \centering
    \includegraphics[width=0.47\textwidth]{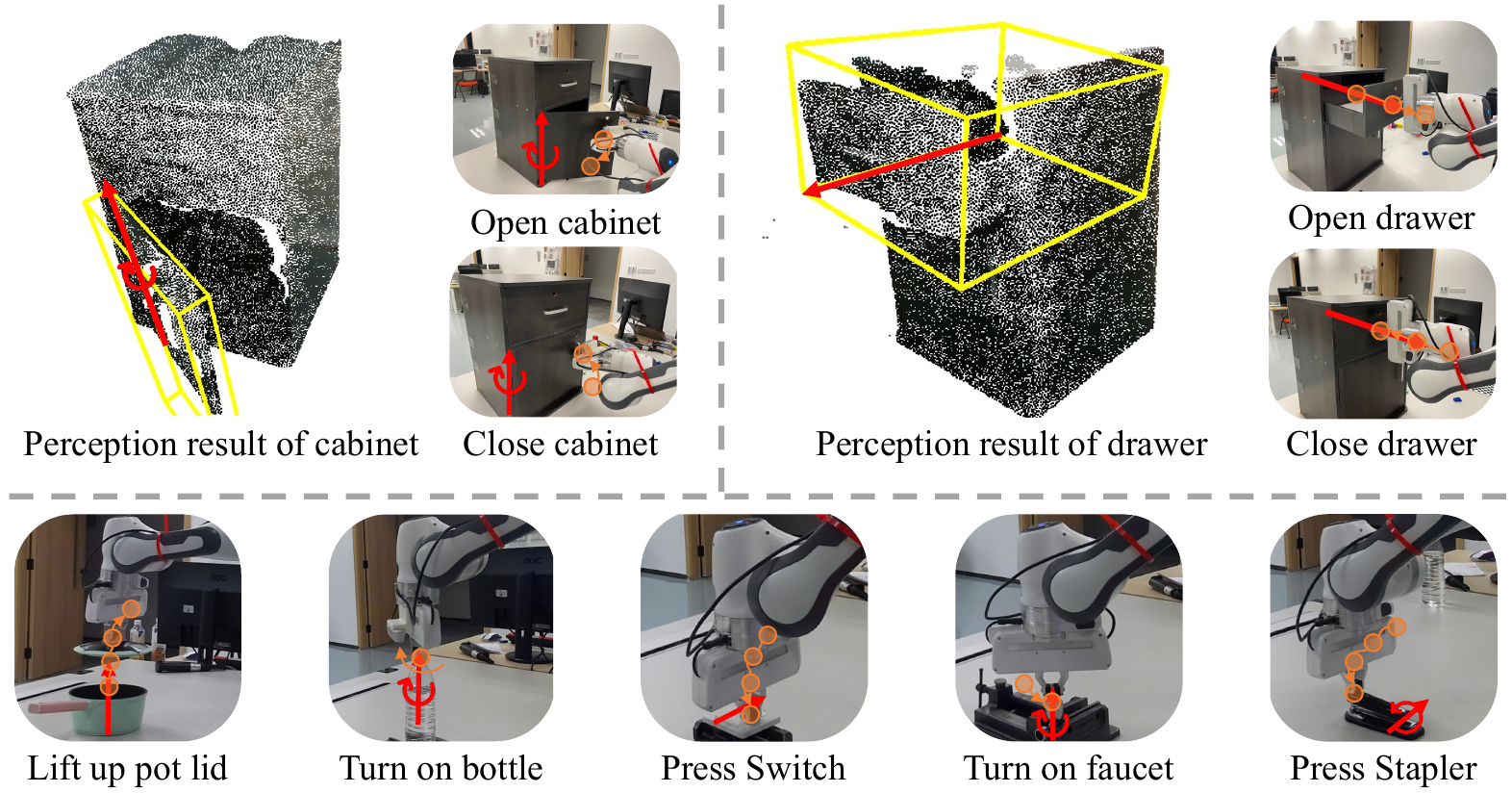}
    
    \caption{Real-world experiments: we generate 3D manipulation waypoints with our framework for real-world object manipulation. The joint information of cabinet and drawer is estimated by the perception model, while the others are provided manually.}
    \label{fig:real}
    \vspace{-1em}
\end{figure}

\subsection{Real-world Experiment}~\label{section:real}
To demonstrate the effectiveness of our framework in practical scenarios, we conducted experiments with a Franka Panda robot arm in the real world. 
To convert the kinematic structure of the manipulation object into texture format with our unified kinematic knowledge parser, we first combine Grounding-DINO~\cite{liu2023grounding} and Segment-anything~\cite{kirillov2023segment} to accurately segment the target object.
We incorporate the GAPartNet~\cite{geng2023gapartnet} as our backbone to detect actionable parts and capture joint information.

As shown in Figure~\ref{fig:real}, we evaluate our framework on 7 distinct objects. For cabinet and drawer categories, we utilize the perception model to obtain the contact point and joint information, while for other categories, we manually provide this information due to the limitation of perception models. Further, we use LLMs to generate 3D manipulation waypoints with our kinematic-aware prompting framework given the joint and affordance information. The results prove that our framework is capable of generalizing to practical scenarios without any additional demonstrations collected in the real world.


\section{CONCLUSIONS and LIMITATIONS}

In this work, we propose a kinematic-aware prompting framework to utilize the rich world knowledge inherent in LLMs for generalizable articulated object manipulation.  Based on the idea that the kinematic structure of an object determines the manipulation policy on it, this framework prompts LLMs with kinematic knowledge of objects to generate low-level motion trajectory waypoints for various object manipulations. Concretely, we build the unified kinematic knowledge parser and kinematic-aware planner, to empower LLMs to understand various object kinematic structures for generalizable articulated object manipulation via in-context learning.
We evaluate our method on 48 instances across 16 categories, and the results prove our method could generalize across unseen instances and categories with only 17 demonstrations for prompting. The real-world experiments also prove our framework's generalization to practical scenarios.

\textbf{Limitations.} Provided with accurate object kinematic knowledge, our framework could achieve generalizable articulated object manipulation. However, its application in the real world is constrained by the capability of existing perception models. It is a promising way to combine the visual foundation models with LLMs for more challenging real-world scenarios. Further, the capability of LLMs in mathematical reasoning and spatial comprehension remains constrained, which would sometimes result in inaccuracy for manipulation waypoint generation. Thus, methods aimed at enhancing mathematical comprehension can be leveraged to augment the efficacy of our object manipulation framework.

\section{acknowledgement}

This research was supported by the Shanghai AI Laboratory, the National Key R\&D Program of China (2022ZD0160100), the National Natural Science Foundation of China (NO.62106272), the Young Elite Scientists Sponsorship Program by CAST (2021QNRC001),  the National Natural Science Foundation of China (62376222), Young Elite Scientists Sponsorship Program by CAST (2023QNRC001) and Public Computing Cloud, Renmin University of China.






\bibliographystyle{IEEEtran}
\bibliography{IEEEabrv,test}


\end{document}